\def\year{2022}\relax
\documentclass[letterpaper]{article} 
\usepackage{aaai21}  
\usepackage{times}  
\usepackage{helvet} 
\usepackage{courier}  
\usepackage[hyphens]{url}  
\usepackage{graphicx} 
\usepackage{natbib}  
\usepackage{caption} 
\usepackage{gensymb}

\usepackage{amsmath}
\usepackage{amssymb}

\title{Accelerating Simulation of Stiff Nonlinear Systems using Continuous-Time Echo State Networks}

\author{
  Ranjan Anantharaman\textsuperscript{\rm 1}, 
  Yingbo Ma\textsuperscript{\rm 2},
  Shashi Gowda\textsuperscript{\rm 1},\\
  Chris Laughman\textsuperscript{\rm 3},
  Viral B. Shah\textsuperscript{\rm 2},
  Alan Edelman\textsuperscript{\rm 1}, 
  Chris Rackauckas\textsuperscript{\rm 1,2}\\
}
\title{Accelerating Simulation of Stiff Nonlinear Systems using Continuous-Time Echo State Networks}

\affiliations{
    \textsuperscript{\rm 1}Massachusetts Institute of Technology\\
    \textsuperscript{\rm 2}Julia Computing\\
    \textsuperscript{\rm 3}Mitsubishi Electric Research Laboratories\\
}

\begin{document}

\maketitle

\begin{abstract}
Modern design, control, and optimization often require multiple expensive simulations of highly nonlinear stiff models. These costs can be amortized by training a cheap surrogate of the full model, which can then be used repeatedly. Here we present a general data-driven method, the continuous-time echo state network (CTESN), for generating surrogates of nonlinear ordinary differential equations with dynamics at widely separated timescales. We empirically demonstrate the ability to accelerate a physically motivated scalable model of a heating system by 98x while maintaining relative error of within 0.2 \%. We showcase the ability for this surrogate to accurately handle highly stiff systems which have been shown to cause training failures with common surrogate methods such as Physics-Informed Neural Networks (PINNs), Long Short Term Memory (LSTM) networks, and discrete echo state networks (ESN). We show that our model captures fast transients as well as slow dynamics, while demonstrating that fixed time step machine learning techniques are unable to adequately capture the multi-rate behavior.  Together this provides compelling evidence for the ability of CTESN surrogates to predict and accelerate highly stiff dynamical systems which are unable to be directly handled by previous scientific machine learning techniques.
\end{abstract}

\section{Introduction}

Stiff nonlinear systems of ordinary differential equations are widely prevalent throughout science and engineering \cite{wanner1996solving,shampine1979user} and are characterized by dynamics with widely separated time scales. These systems require highly stable numerical methods to use non-vanishing step-sizes reliably \cite{gear1971numerical}, and also tend to be computationally expensive to solve. Even with state-of-the-art simulation techniques, design, control, and optimisation of these systems remains intractable in many realistic engineering applications \cite{benner2015survey}. To address these challenges, researchers have focused on techniques to obtain an approximation to a system (called a ``surrogate") whose forward simulation time is relatively inexpensive while maintaining reasonable accuracy \cite{willard2020integrating, ratnaswamy2019physics, zhang2020hydrological, kim2020fast, vanfast}.  


\begin{figure*}
    \centering
    \includegraphics[scale = 0.67]{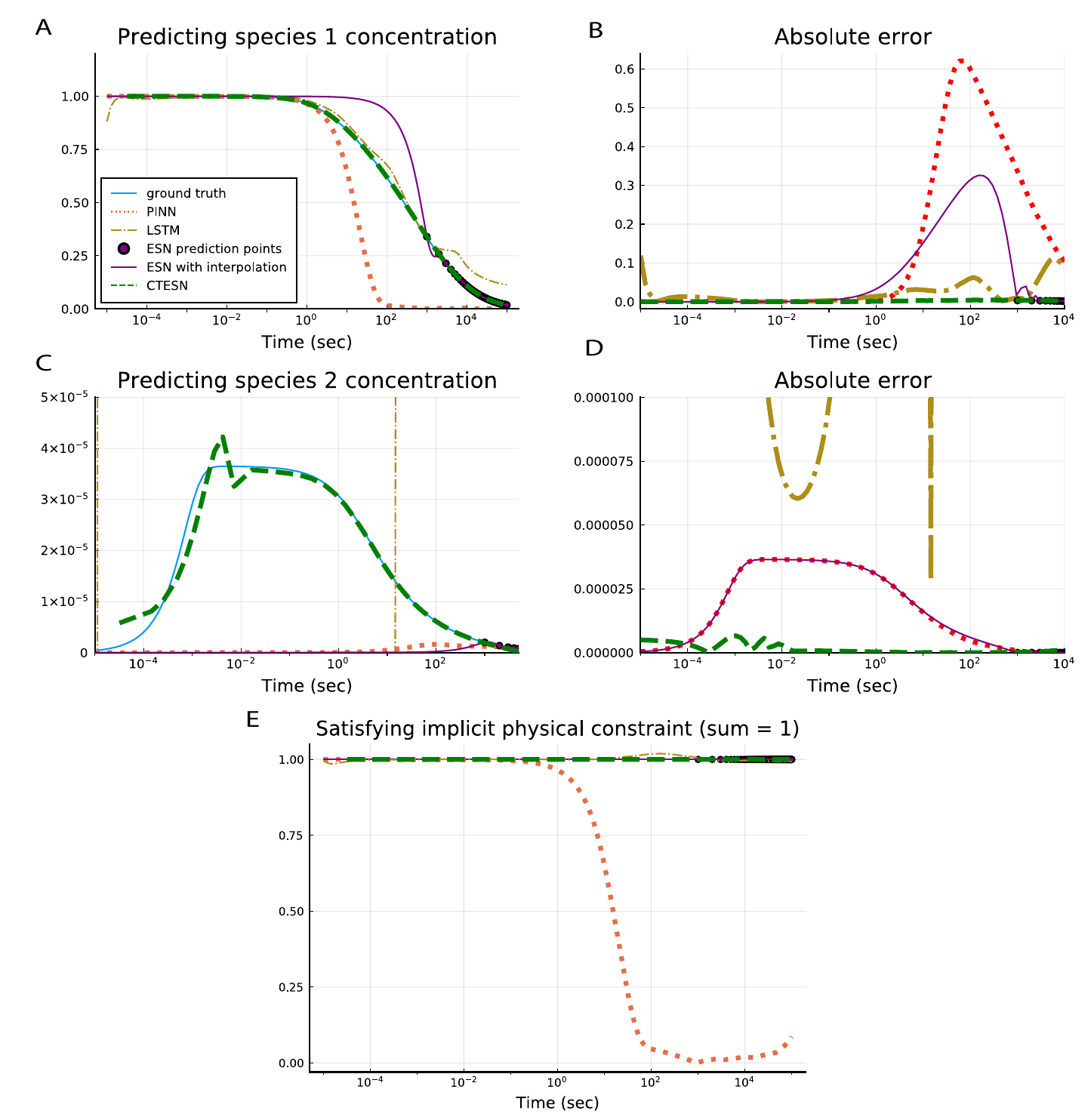}
    \caption{\textbf{Prediction of each surrogate on the Robertson's equations} Shown in each figure is the result of the data-driven algorithm's prediction at $p =[0.04, 3 \times 10^7,  1\times 10^4]$, a parameter set not in the training data. Ground truth, obtained by solving the ODE using the Rosenbrock23 solver with absolute tolerance of $10^{-6}$, is in blue. The PINN was trained using a 3-layer multi-layer perceptron with the ADAM optimizer for 300,000 epochs with minibatching, and its prediction is in red. Both the ESN and CTESN were trained with a reservoir size of 3000 on  a parameter space of $[0.036, 0.044] \times [2.7 \times 10^7, 3.3 \times 10^7] \times [9\times 10^3, 1.1 \times 10^4]$, from which 1000 sets of parameters were sampled using Sobol sampling. The predictions of the CTSEN are generated by the radial basis function prediction of $W_{out}(p)$ and are shown in green. Predictions from the ESN are in purple. The LSTM predictions, in gold, are generated by a network with 3 hidden LSTM layers and an output dense layer, after training for 2000 epochs. (A) A timeseries plot of the $y_1(t)$ predictions. (B) The absolute error of the surrogate predictions on $y_1(t)$. (C) A timeseries plot of the $y_2(t)$ predictions. (D) The absolute error of the surrogate predictions on $y_2(t)$. (E) The result of $y_1(t)+y_2(t)+y_3(t)$ over time. By construction this quantity's theoretical value is 1 over the timeseries.}
    \label{fig:rober}
\end{figure*}


A popular class of traditional surrogatization techniques is projection based model order reduction, such as the proper orthogonal decomposition (POD) \cite{benner2015survey}. This method computes ``snapshots" of the trajectory and uses the singular value decomposition of the linearization in order to construct a basis of a subspace of the snapshot space, and the model is remade with a change of basis. However, if the system is very nonlinear, the computational complexity of this linearization-based reduced model can be almost as high as the original model. One way to overcome this difficulty is through empirical interpolation methods \cite{nguyen2014model}.  Other methods to produce surrogates generally utilize the structural information known about highly regular systems like partial differential equation discretizations \cite{frangos2010surrogate}. 


Many of these methods require a scientist to actively make choices about the approximations being performed to the system. In contrast, the data-driven approaches like Physics-Informed Neural Networks (PINNs)\cite{raissi2019physics} and Long Short Term Memory (LSTM) networks \cite{chattopadhyay2020data} have gained popularity due to their apparent applicability to ``all'' ordinary and partial differential equations in a single automated form. However, numerical stiffness \cite{soderlind2015stiffness} and multiscale dynamics represent an additional challenge. Highly stiff differential equations can lead to gradient pathologies that make common surrogate techniques like PINNs hard to train \cite{wang2020understanding}. 

A classic way to create surrogates for stiff systems is to simply eliminate the stiffness. The computational singular perturbation (CSP) method \cite{hadjinicolaou1998asymptotic} has been shown to decompose chemical reaction equations into fast and slow modes. The fast modes are then eliminated, resulting in a non-stiff system. Another option is to perform problem-specific variable transformations \cite{qian2020lift, kramer2019nonlinear} to a form more suited to model order reduction by traditional methods. These transformations are often problem specific and require a scientist's intervention at the equation level. Recent studies on PINNs have demonstrated that such variable elimination may be required to handle highly stiff equations because the stiffness leads to ill-conditioned optimization problems. For example, on the classic Robertson's equations (ROBER) \cite{robertson1976numerical} and Pollution model (POLLU) \cite{verwer1994gauss} stiff test problems it was shown that direct training of PINNs failed, requiring the authors to perform a quasi-steady state (QSS) assumption in order for accurate prediction to occur \cite{ji2020stiff}. However, many chemical reaction systems require transient activations to properly arrive at the overarching dynamics, making the QSS assumption only applicable to a subset of dynamical systems \cite{henry2016short, flach2006use, eilertsen2020quasi, turanyi1993error, schuster1989generalization, thomas2011communication}. Thus while demonstrating promising results on difficult equations, training on the QSS-approximated equations requires specific chemical reaction networks and requires the scientist to make approximation choices that are difficult to automate, which reduces the general applicability that PINNs were meant to give.

The purpose of this work is to introduce a general data-driven method, the CTESN, that is generally applicable and able to accurately capture highly nonlinear heterogeneous stiff time-dependent systems without requiring the user to train on non-stiff approximations. It is able to accurately train and predict on highly ill-conditioned models. We demonstrate these results (Figure \ref{fig:rober}) on the Roberston's equations, which PINNs, LSTM networks and discrete-time machine learning techniques fail to handle. Our results showcase the ability to transform difficult stiff equations into non-stiff reservoir equations which are then integrated in place of the original system. Given the $O(n^3)$ scaling behavior of general stiff ODE solvers due to internal LU-factorizations, the resulting approximation by a surrogate with linear scaling with number of outputs, we observe increasing accelerations as the system gets larger. With this scaling difference we demonstrate the ability to accelerate a large stiff system by 98x while achieving $<0.2\%$ error (Figure \ref{fig:bench}). 
 

\section{Continuous-Time Echo State Networks}


Echo State Networks (ESNs) are a reservoir computing framework which projects signals from higher dimensional spaces defined by the dynamics of a fixed non-linear system called a ``reservoir" \cite{ozturk2007analysis}. The ESN's mathematical formulation is as follows. For a $N_R$-dimensional reservoir, the reservoir equation is given by:
\begin{equation}
r_{n+1} = f(A r_n + W_{fb} x_n),
\end{equation}
where $f$ is a chosen activation function (like tanh or sigmoid), $A$ is the $N_R \times N_R$ reservoir weight matrix, and $W_{fb}$ is the $N_R \times N$ feedback matrix where $N$ is the size of our original model. In order to arrive at a prediction of our original model, we take a projection of the reservoir:
\begin{equation}
\hat{x}_n = g(W_{out} r_n),
\end{equation}
where $g$ is the output activation function (generally the identity or sigmoid) and $W_{out}$ is the $N \times N_R$ projection matrix. In the training process of an ESN, the matrices $A$ and $W_{fb}$ are randomly chosen constants, meaning the $W_{out}$ matrix is the only variable which needs to be approximated. $W_{out}$ is calculated by using a least squares fit of against the model's time series, which then fully describes the prediction process.

This process of using a direct linear solve, such as a QR-factorization, to calculate $W_{out}$ means that no gradient-based optimization is used in the training process. For this reason ESNs have traditionally been used as a stand-in for recurrent neural networks which overcome the vanishing gradient problem \cite{jaeger2007optimization, lukovsevivcius2009reservoir, mattheakis2019recurrent, vlachas2020backpropagation, chattopadhyay2019data, grezes2014reservoir, evanusa2020deep, butcher2013reservoir}. However, ESNs have also been applied to learning chaotic systems \cite{chattopadhyay2019data, doan2019physics}, nonlinear systems identification \cite{jaeger2003adaptive}, bio-signal processing \cite{kudithipudi2016design}, and robot control \cite{polydoros2015advantages}. These are all cases where the derivative calculations are unstable or, as in the case of chaotic equations, are not well-defined for long time spans.

This ability to handle problems with gradient pathologies gives the intuitive justification for exploring reservoir computing techniques on handling stiff equations. However, stiff systems generally have behavior spanning multiple timescales which are difficult to represent with uniformly-spaced prediction intervals. For example, in the ROBER problem we will showcase, an important transient occurs for less than a 10 seconds of the 10,000 second simulation. However this feature is important to capture the catalysis that kick-starts the long-term changes. Many more samples from $t\in[0,10]$ will be required than from $t\in[10,10^5]$ in order to accurately capture the dynamics of the system. These behaviors are the reason why all of the major software for handling stiff equations, such as CVODE \cite{hindmarsh2005sundials}, LSODA \cite{hindmarsh2005lsoda}, and Radau \cite{hairer1999stiff} are adaptive. In fact, this behavior is so crucial to the stable handling of stiff systems that robust implicit solves tie the stepping behavior to the implicit handling of the system with complex procedures for reducing time steps when Newton convergence rates are reduced \cite{wanner1996solving, hosea1996analysis, hairer1999stiff}. For these reasons, we will demonstrate that the classic fixed time step reservoir computing methods from machine learning are unable to handle these highly stiff equations.

To solve these issues, we introduce a new variant of ESNs, which we call continuous-time echo state networks (CTESNs), which allows for an underlying adaptive time process while avoiding gradient pathologies in training. Let $N$ be the dimension of our model, and let $P$ be a Cartesian space of parameters under which the model is expected to operate. The CTESN of with reservoir dimension $N_R$ is defined as
\begin{align}
    r^\prime &= f(Ar + W_{hyb} x(p^*,t)),\\
    x(t) &= g(W_{out} r(t)),
\end{align}
where $A$ is a fixed sparse random matrix of dimension $N_R \times N_R$ and $W_{hyb}$ is a fixed random dense matrix of dimensions $N_R \times N$. The term $W_{hyb}x(p^*, t)$ represents a ``hybrid'' term that incorporates physics information into the reservoir \cite{pathak2018hybrid}, namely a solution at some point in the parameter space of the dynamical system. Given these fixed values, the readout matrix $W_{out}$ is the only trained portion of this network and is obtained through a least squares fit of the reservoir ODE solution against the original timeseries. We note that in this study we choose $f=\text{tanh}$ and $g=\text{id}$ for all of our examples.

To obtain a surrogate that predicts the dynamics at new physical parameters, the reservoir projection $W_{out}$ is fit against many solutions at parameters $\{p_1, \dots, p_n\}$, where $n$ is the number of training data points sampled from the parameter space. Using these fits, an interpolating function $W_{out}(p)$ between the matrices can be trained. A prediction $\hat{x}(t)$ for at physical parameters $\hat{p}$ is thus given by:
\begin{equation}
\hat{x}(t) = W_{out}(\hat{p})  r(t).
\end{equation}

A strong advantage of our method is its ease of implementation and ease of training. Global $L_2$ fitting via stabilized methods like SVD are robust to ill-conditioning, alleviating many of the issues encountered when attempting to build neural surrogates of such equations. Also note that in this particular case, the readout matrix is fit against the same reservoir time series. This means that prediction does not need to simulate the reservoir, providing an extra acceleration. 

Another advantage is the ability to use time stepping information from the solver during training. As noted before, not only are step sizes chosen adaptively based on minimizing a local error estimate to a specified tolerance \cite{shampine1979user}, but they also adapt to concentrate around the most stiff and numerically difficult time points of the model by incorporating the Newton convergence into the rejection framework. These timestamps thus provide heuristic snapshots of the most important points for training the least squares fit, whereas snapshots from uniform time steps may skip over many crucial aspects of the dynamics.

\section{Training}
In this section we describe the automated training procedure used to generate CTESN surrogates. An input parameter space $P$ is first chosen. This could be a design space for the model or a range of operating conditions. Now $n$ sets of parameters $\{p_1, \dots , p_n\}$ are sampled from this space using a sampling sequence that covers as much of the space as possible. The full model is now simulated at each sample in parallel since each run is independent, generating time series for each run. The choice of points in time used to generate the time series at each $p$ comes from the numerical ODE solve at that $p$. The reservoir ODE is then constructed using a candidate solution at any one of the $n$ parameters $x(p^*,t), p^* \in \{p_1, \dots , p_n\}$ and is then simulated, generating the reservoir time series. Since the reservoir ODE is non-stiff, this simulation is cheap compared to the cost of the full model. Least squares projections can now be calculated from each solution to the reservoir in parallel. Once all the least squares matrices are obtained, an interpolating function is trained to predict the least squares projection matrix. Both the least squares fitting and training the interpolating function are, in practice, much cheaper than the cost of simulating the model multiple times.

Prediction comprises of two steps: predicting the least squares matrix, and simulating the reservoir time series (or, in this case, just using the pre-computed continuous solution since the reservoir is fixed for every set of parameters). The final prediction is just the matrix multiplication of two.

A strong advantage of the training is that it requires no manual investigation of the stiff model on the part of the researcher and can be called as an off-the-shelf routine. It allows the researcher to make a trade-off, computing a few parallelized runs of the full stiff model in order to generate a surrogate, which can then be plugged in and used repeatedly for design and optimization.

We implemented the training routines and the following models in the Julia programming language \cite{bezanson2017julia} to take advantage of its first class support for differential equations solvers \cite{rackauckas2017differentialequations} and scientific machine learning packages. For the examples in this paper, we have sampled the high-dimensional spaces using a Sobol low-discrepancy sequence \cite{sobol2011construction} and interpolated the $W_{out}$ matrices using a radial basis function provided by the Julia package Surrogates.jl (\texttt{https://github.com/SciML/Surrogates.jl}).

\section{Case Studies}
In this section we describe two representative examples. We demonstrate that the CTESN can handle highly stiff behaviour through the ROBER example. We then talk about the performance of the surrogate on a scalable, physically-inspired heating system.

\subsection{Robertson Equations and High Stiffness}

We first consider Robertson's chemical reactions involving three reaction species $A$, $B$ and $C$:

\begin{align*}
    A  &\xrightarrow[]{0.04} B \\
    B + B &\xrightarrow[]{3\times 10^7} C + B \\
    B + C &\xrightarrow[]{10^4} A + C
\end{align*}

\noindent which lead to the ordinary differential equations:

\begin{align} \label{eq:robert}
\dot{y_1} &= -0.04 y_1+10^{4} y_2 \cdot y_3 \\
\dot{y_2} &= 0.04 y_1-10^{4} y_2 \cdot y_3-3 \cdot 10^{7} y_2^{2} \\
\dot{y_3} &= 3 \cdot 10^{7} y_2^{2}
\end{align}

where $y_1$, $y_2$, and $y_3$ are the concentrations of $A$, $B$ and $C$ respectively. This system has widely separated reaction rates ($0.04, 10^{4}, 3\cdot10^{7}$), and is well known to be very stiff \cite{gobbert1996robertson, robertson1975some, robertson1976numerical}. It is commonly used as an example to evaluate integrators of stiff ODEs \cite{hosea1996analysis}. Finding an accurate surrogate for this system is difficult because it needs to capture both the stable slow reacting system and the fast transients. Additionally, the surrogate needs to be consistent with this system's implicit physical constraints, such as the conservation of matter ($y_1 + y_2 + y_3 = 1$) and positivity of the variables ($y_i > 0$), in order to provide a stable solution.  

A surrogate was trained by sampling 1000 sets of parameters from the Cartesian parameter space $[0.036, 0.044] \times [2.7 \times 10^7, 3.3 \times 10^7] \times [9\times 10^3, 1.1 \times 10^4]$ using Sobol sampling so as to evenly cover the whole space. We train on the time series of the three states themselves as outputs. A least squares projection $W_{out}$ was fit for each set of parameters, and then a radial basis function was used to interpolate between the matrices. The prediction workflow is as follows: given a set of parameters whose time series is desired, the radial basis function predicts the projection matrix. The pre-simulated reservoir is then sampled at the desired time points, and a matrix multiplication with the predicted $W_{out}$ gives us the desired prediction. Figure \ref{fig:rober} shows a comparison between the CTESN, ESN, PINN and LSTM methods. The PINN data is reproduced from \cite{ji2020stiff} and the ESN was trained using 101 time points uniformly sampled from the time span, while CTESN used 61 adaptively sampled time points informed by the ODE solver (Rosenbrock23 \cite{shampine1982implementation}). 

The CTESN method is able to accurately capture both the slow and fast transients and respect the conservation of mass. The ESN is able to accurately predict at the time points it was trained on, but many features are missed. The uniform stepping completely misses the fast transient rise at $t = 10^{-4}$ because the uniform intervals do not sample points from that time scale. Additionally, the first sampled time point at $t=100$ is far into the concentration drop of $y_1$ which leads to an inaccurate prediction before the system settles into near steady state behavior. As stated earlier, the CTESN uses information from a stiff ODE solver to choose the right points along the time span to accurately capture multi-scale behaviour with less training data than the ESN. In order to compare the discrete ESN to the continuous result, a cubic spline was fit to its 101 evenly spaced prediction points.

\begin{figure}
    \centering
    \includegraphics[scale = 0.4]{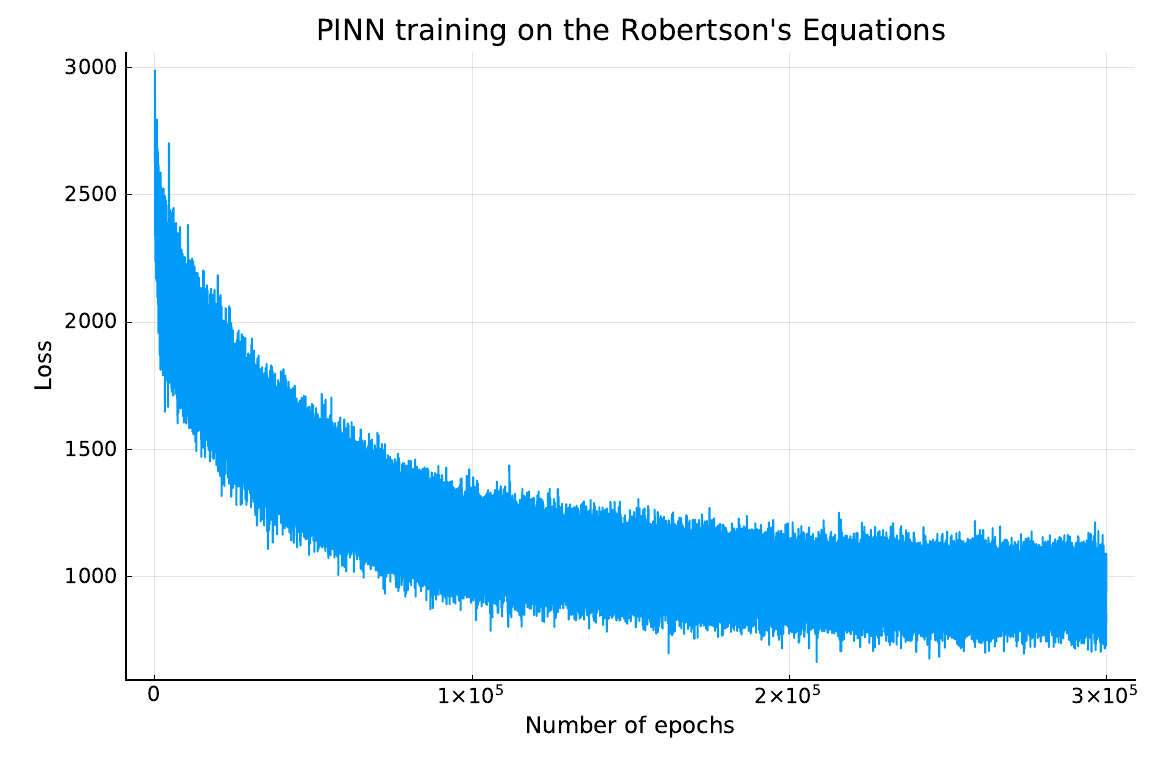}
    \caption{\textbf{Training a PINN on the Robertson's Equations}: PINN was trained for 300,000 epochs using the ADAM optimizer with a learning rate of $10^{-3}$, by which time the loss seems to saturate. The hyperparameters of the PINN can be found in the Case Studies section.}
    \label{fig:pinntrain}
\end{figure}

The PINN was trained by sampling 2500 logarithmically spaced points across the time span. The network used was a 3-layer multi-layer perceptron with 128 nodes per hidden layer and the Gaussian Error Linear Unit activation function \cite{hendrycks2016gaussian}. The layers were initialed using Xavier initialization \cite{glorot2010understanding}, and trained with the ADAM optimizer \cite{kingma2019method}  at a learning rate of $10^{-3}$ for 300,000 epochs with mini batch size of 128. Figure \ref{fig:pinntrain} shows the convergence plot as the PINN trains on the ROBER equations. The LSTM network used a similar architecture to the  PINN, but with LSTM hidden layers instead of fully connected layers. It used 2500 logarithmically spaced points and was trained for 2000 epochs until convergence.  

Figure \ref{fig:rober} highlights how the trained PINN fails to capture both the fast and the slow transients and do not respect mass conservation. Our collaborators investigated why PINNs fail to solve these equations in \cite{ji2020stiff}. The reason for the difficulty can be attributed to recently identified results in gradient pathologies in the training arising from stiffness \cite{wang2020understanding}. With a highly ill-conditioned Hessian in the training process due to the stiffness of the equation, it is very unlikely for local optimization to find a parameters which make an accurate prediction. We additionally note stiff systems of this form may be hard to capture by neural networks directly as neural networks show a bias towards low frequency functions \cite{rahaman2019spectral}.

\begin{figure}
    \centering
    \includegraphics[scale = 0.5]{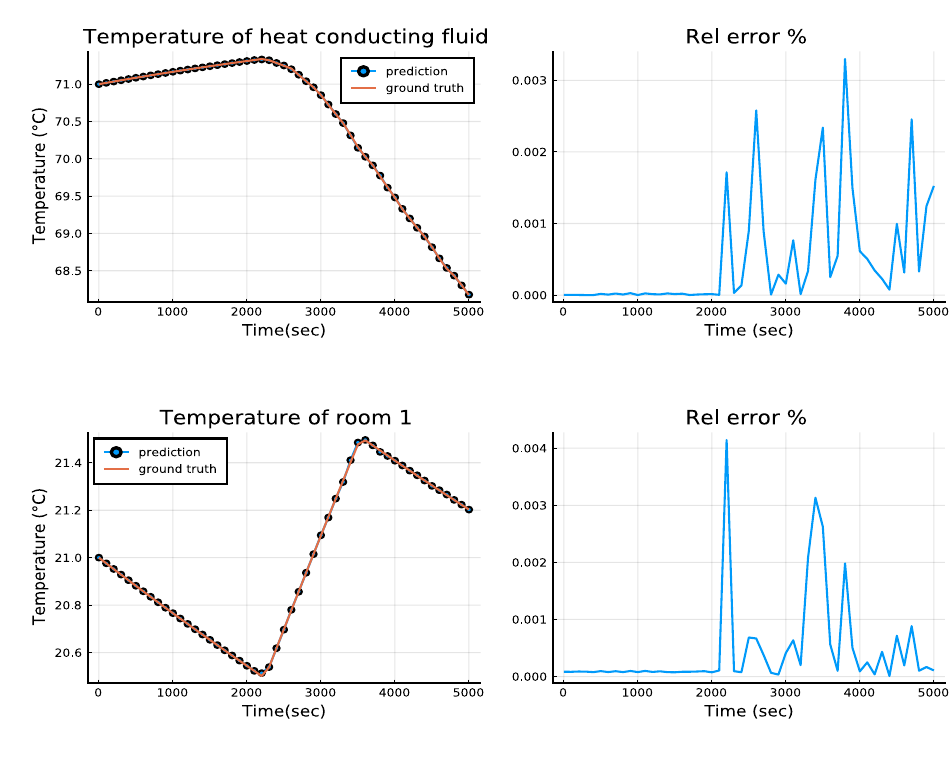}
    \caption{\textbf{Validating the surrogate of the scalable heating system with 10 rooms.} When tested with parameters it has not seen in training, our surrogate is able to reproduce the behaviour of the system to within 0.01\% error. The surrogate is trained on 100 points sampled from the $[17\degree C, 23\degree C] \times [65\degree C, 75\degree C]$ where the ranges represent set point temperature of each room and set point of the fluid supplying heat to the rooms respectively. The test parameters that validated here are $[21\degree C, 71\degree C]$. More details on training can be found in the Case Studies section.}
    \label{fig:heating}
\end{figure}

\subsection{Stiffly-Aware Surrogates of HVAC Systems}

\begin{figure}
    \centering
    \includegraphics[scale = 0.4]{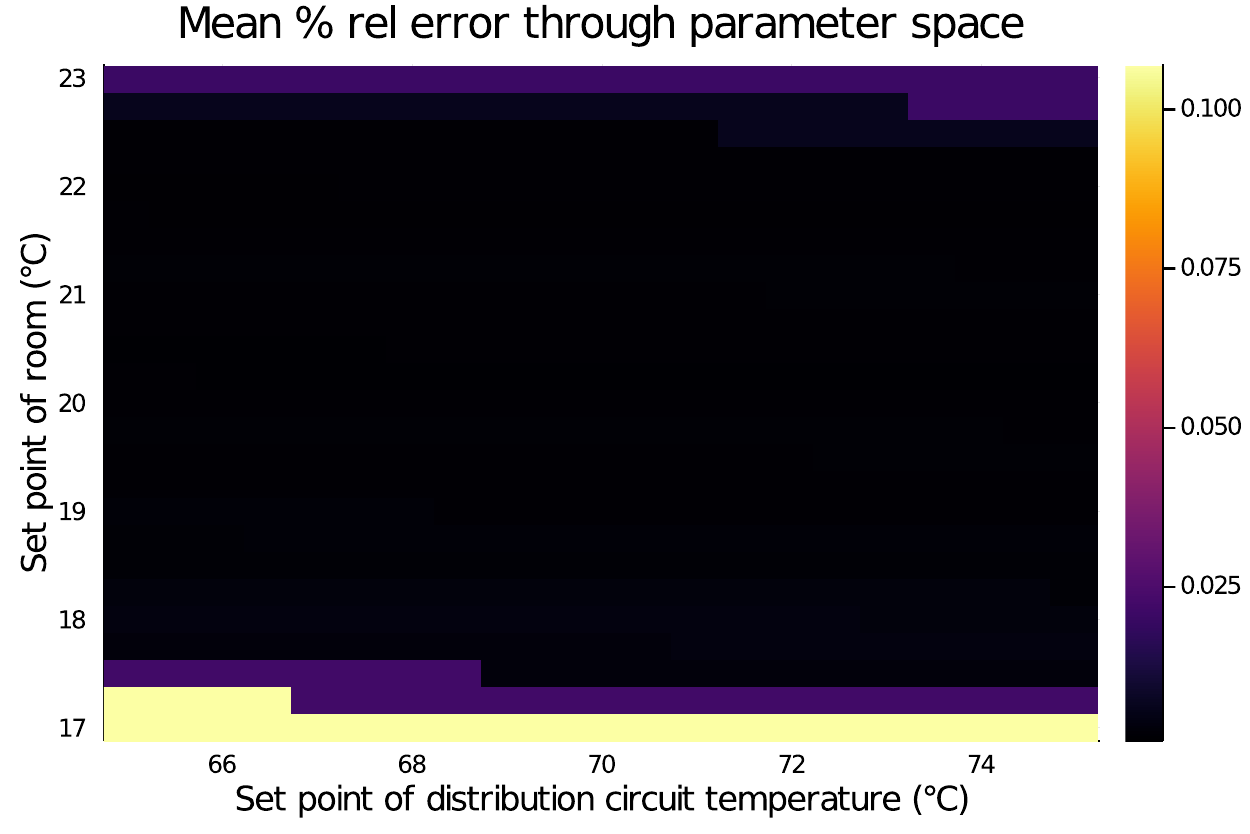}
    \caption{\textbf{Reliability of surrogate through parameter space.} We sampled our grid at over 500 grid points and plotted a heatmap of test error through our parameter space. We find our surrogate performs reliably even at the border of our space with error within 0.1\% }
    \label{fig:heatmap}
\end{figure}

Our second test problem is a scalable benchmark used in the engineering community \cite{casella2015modelica}. It is a simplified, lumped-parameter model of a heating system with a central heater supplying heat to several rooms through a distribution network. Each room has an on-off controller with hysteresis which provides very fast localized action \cite{ranade2014multi}. The resulting system of equations is thus very stiff and unable to be solved by standard explicit time stepping methods. 

The size of the heating system is scaled by a parameter $N$ which refers to the number of users/rooms. Each room is governed by two equations corresponding to its temperature and the state of its on-off controller. The temperature of fluid supplying heat to each room is governed by one equation. This produces a system with $2N + 1$ coupled non-linear equations. This ``scalability'' lets us test how our CTESN surrogate scales. To train the surrogate, we define a parameter space $P$ under which we expect it to operate. First, we assume set point temperature of each room to be between $17 \degree C$ and $23 \degree C$. Each room is warmed by a heat conducting fluid, whose set point is between $ 65\degree C$ and $75\degree C$. Thus the parameter space over which we expect our surrogate to work is the rectangular space denoted by $[17\degree C, 23\degree C] \times [65\degree C, 75\degree C]$. 

\begin{figure}
    \centering
    \includegraphics[scale = 0.4]{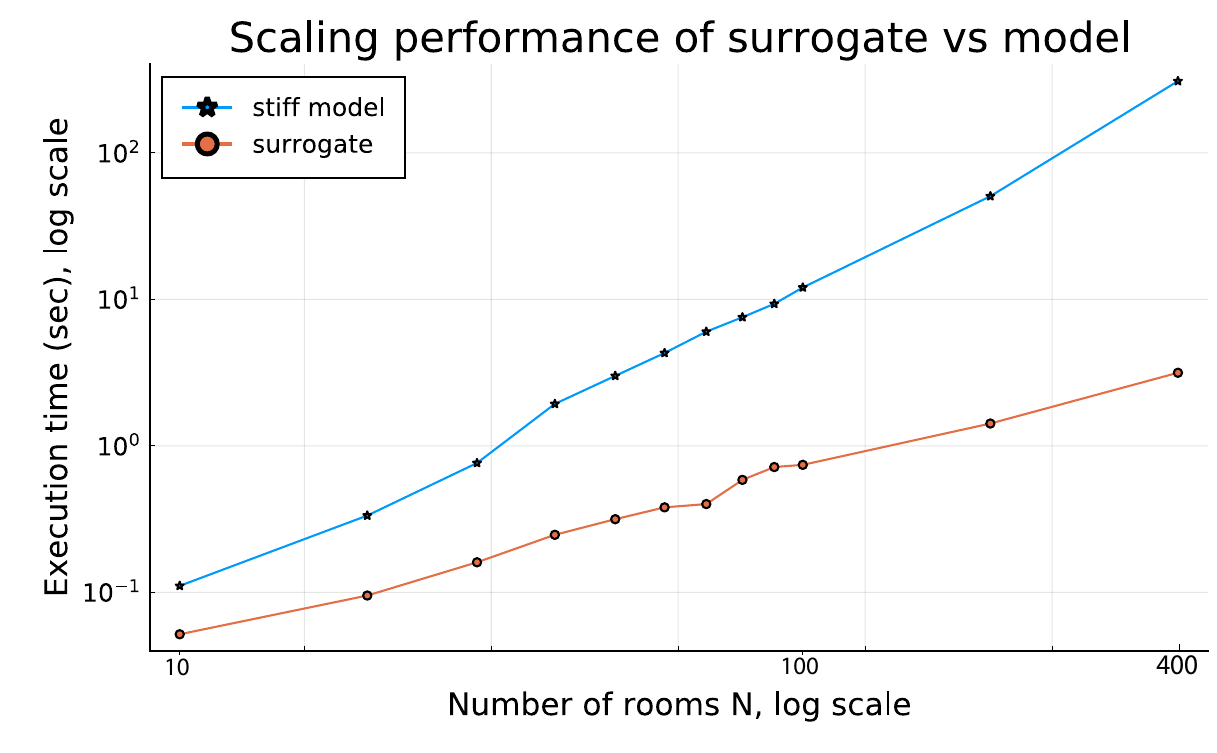}
    \caption{\textbf{Scaling performance of surrogate on heating system.} We compare the time taken to simulate the full stiff model to the trained surrogate with 10, 20, 30 , 40, 50, 60, 70, 80, 90, 100, 200 and 400 rooms. We observe a speedup of up to 98x. The surrogate was trained by sampling 100 sets of parameters from our input space, with a reservoir size of 3000.}
    \label{fig:bench}
\end{figure}

We used a reservoir size of 3000 and sampled 100 sets of parameters from this space using Sobol sampling, and fit least squares projection matrices $W_{out}$ between each solution and the reservoir. For a system with $N$ rooms, we train on $N+1$ outputs, namely the temperature of each room, and the temperature of the heat conducting fluid. Figure \ref{fig:heating} demonstrates that the training technique is accurately able to find matrices $W_{out}$ which capture the stiff system within $0.01\%$ error on a test parameters. We then fit an interpolating radial basis function $W_{out}(p)$. Figure \ref{fig:heatmap} demonstrates that the interpolated $W_{out}(p)$ is able to adequately capture the dynamics throughout the trained parameter space. Lastly, Figure \ref{fig:bench} demonstrates the $O(N)$ cost of the surrogate evaluation, which in comparison to the $O(N^3)$ cost of a general implicit ODE solver (due to the LU-factorizations) leads to an increasing gap in the solver performance as $N$ increases. At the high end of our test, the surrogate accelerates a 801 dimensional stiff ODE system by approximately 98x.

\section{Conclusion \& Future Work}
We present CTESNs, a data-driven method for generating surrogates of nonlinear ordinary differential equations with dynamics at widely separated timescales. Our method maintains accuracy for different parameters in a chosen parameter space, and shows favourable scaling with system size on a physics-inspired scalable model. This method can be applied to any ordinary differential equation without requiring the scientist to simplify the model before surrogate application, greatly improving productivity.


In future work, we plan to extend the formulation to take in forcing functions.This entails that the reservoir needs to be simulated every single time a prediction is made, adding to running time, but we do note that numerically simulating the reservoir is quite fast in practice as it is non-stiff, and thus techniques which regenerate reservoirs on demand will likely not incur a major run time performance cost. 

Our method utilizes the continuous nature of differential equation solutions. Hybrid dynamical systems, such as those with event handling \cite{ellison1981efficient}, can introduce discontinuities into the system which will require extensions to our method. Further extensions to the method will handle both derivative discontinuities and events present in Filippov dynamical systems \cite{filippov2013differential}.Further opportunities could explore utilizing more structure within equations for building a more robust CTESN or decrease the necessary size of the reservoir.

To train both the example problems in this paper, we required no knowledge of the physics. This presents an opportunity to train surrogates of black-box systems.


\section*{Acknowledgement}

The information, data, or work presented herein was funded
in part by the Advanced Research Projects Agency-Energy (ARPA-E), U.S.
Department of Energy, under Award Numbers DE-AR0001222 and DE-AR0001211, and NSF awards OAC-1835443 and IIP-1938400. The views and
opinions of authors expressed herein do not necessarily state or reflect those of
the United States Government or any agency thereof. The authors thank Francesco Martinuzzi for reviewing drafts of this paper. 

\bibliography{references.bib}

\end{document}